%% file: neurips_2026.tex
\documentclass{article}

\PassOptionsToPackage{numbers}{natbib}

 \usepackage[main, final]{neurips_2026}

\usepackage[utf8]{inputenc}
\usepackage[T1]{fontenc}
\usepackage{hyperref}
\usepackage{url}
\usepackage{booktabs}
\usepackage{amsfonts}
\usepackage{nicefrac}
\usepackage{microtype}
\usepackage{xcolor}
\definecolor{citeblue}{HTML}{2A78D6}
\definecolor{refblue}{HTML}{114290}
\definecolor{urlblue}{HTML}{4D8FE6}
\hypersetup{colorlinks=true, citecolor=citeblue, linkcolor=refblue, urlcolor=urlblue}

\usepackage{amsmath}
\usepackage{enumitem}
\newcommand{\vx}{x}
\newcommand{\vv}{v}
\newcommand{\ystar}{y^{\star}}
\newcommand{\yhat}{\hat{y}}
\newcommand{\pS}{p_{S}}
\newcommand{\pT}{p_{T}}
\newcommand{\JSD}{\mathrm{JSD}}
\newcommand{\KL}{\mathrm{KL}}
\newcommand{\sg}{\mathrm{sg}}
\newcommand{\E}{\mathbb{E}}
\usepackage{booktabs}
\usepackage{array}
\usepackage{xcolor}
\usepackage{colortbl}
\usepackage{graphicx}

\title{Watching Itself Watch: \\ Self-Auditing Visual Reliance for Video Reasoning
}

\author{
Rohan Dugad\space\space\space 
Mi Luo\space\space\space 
Kristen Grauman\space\space\space
\vspace{1mm} \\
The University of Texas at Austin\qquad \\
}

\input{preamble}
\begin{document}

\maketitle

\begin{abstract}
Video reasoning hinges on a small number of visually decisive tokens embedded in otherwise largely language-driven chains of thought. Yet prevailing post-training methods such as GRPO overlook this structure, assigning coarse trajectory-level rewards while requiring multiple rollouts and often costly external judges. We introduce \textbf{Vi}sually-Aware \textbf{O}n-Policy \textbf{S}elf-Distillation (VIOS), a post-training framework that identifies where a model under-utilizes visual evidence and concentrates learning at those positions. Our key insight is that a model can \emph{audit its own visual reliance}: the divergence between its next-token distributions with and without the video reveals how strongly each prediction depends on the visual input. Measuring this reliance for both a student and an answer-privileged teacher exposes a per-token \emph{reliance gap}, identifying positions where the teacher exploits visual evidence that the student overlooks. VIOS translates this signal into dense, targeted supervision through three complementary components: an exponential moving average (EMA) teacher that co-evolves with the student, a contrastive objective that encourages predictions to depend on the video, and a reliance-gap gate that focuses learning on visually under-utilized tokens. Crucially, VIOS requires only a single on-policy rollout from a single model, without external rewards, judges, or larger teachers. Across 12 benchmarks spanning general, temporal, spatial, knowledge-intensive, and long-video understanding, VIOS consistently improves three model backbones and outperforms contemporary video-reasoning systems, while training in over $9\times$ fewer steps, $60\times$ fewer FLOPs, and $111\times$ fewer generated tokens.\footnote{Project page: \textcolor{blue}{\url{https://vision.cs.utexas.edu/projects/VIOS/}}}
\end{abstract}

\input{1-introduction}

\input{2-related-work}

\input{3-method}

\input{4-experiments}

\input{5-conclusion}

{
\small
\addtolength{\bibsep}{-1.5pt}
\bibliographystyle{plainnat}
\bibliography{main}
}

\newpage

\appendix

\input{appendix}

\end{document}

%% file: preamble.tex
\usepackage{caption}

%% file: 1-introduction.tex
\section{Introduction}

Consider a video in which deciding whether one event caused another rests on a brief transition across two frames. The narration around that transition does not describe it and the prior context does not imply it. Video reasoning in multimodal large language models (MLLMs) is characterized by exactly this kind of dependency~\citep{video_r1, videorft, video_r2}, where a correct continuation hinges on specific moments of visual evidence that no textual fluency can substitute for. A model must track entities through occlusion, relate causes to outcomes that are mere seconds apart, and revise its working hypothesis as new frames arrive~\citep{video_survey, crossvid, mars2}. What distinguishes this regime from language-only reasoning is not the length of the chain of thought, but the fact that particular positions in that chain are answerable only by careful attention to the visual input.

Despite the importance of visually decisive moments, existing video post-training methods provide coarse supervision at substantial computational cost. Supervised fine-tuning is off-policy, trains the student on prefixes it never produces itself, and tends to memorize surface patterns of the demonstrations~\citep{agarwal_opd, opsd, sft_memorize, anchored_sft}. Group Relative Policy Optimization (GRPO)~\citep{deepseekmath, deepseek_r1} closes the on-policy gap by sampling multiple rollouts per prompt and rewarding those that score above the group mean while penalizing those below it. This estimator typically requires eight or more rollouts per prompt~\citep{deepseekmath, deepseek_r1, dapo}, wastes gradient on groups that collapse to identical rewards~\citep{dapo, ngrpo, agpo}, and applies a single scalar reward across the trajectory even though only a small minority of tokens determine its outcome~\citep{forking_tokens, tempo, causal_credit}. Recent video methods compound this cost by relying on substantially larger judge models to assign rewards online~\citep{videochat_r1, video_r2, video_ver}. Ultimately, these methods remain costly while failing to identify where the model under-utilizes visual evidence during reasoning. 

Our central observation is that this missing structure can be recovered from the model itself. We introduce \textbf{Vi}sually-Aware \textbf{O}n-Policy \textbf{S}elf-Distillation (\textbf{VIOS}), a novel post-training procedure that recovers dense, per-token, vision-aware supervision from a single rollout and a single model. 

Our framework consists of a student and a teacher, both initialized from the same base model. Both receive the video and question, while the teacher additionally sees the ground-truth answer. The student generates an on-policy rollout, and the teacher provides next-token distributions for the student to match at each position. The teacher is updated through an exponential moving average (EMA) of the student, yielding a co-evolving supervision target. However, matching the teacher alone does not ensure that the student actually uses the video. To address this, we measure the student's \emph{visual reliance} at each token by comparing its next-token distributions with and without the video in context. A larger difference indicates that the prediction is more sensitive to visual input. We introduce a contrastive objective that encourages this sensitivity, pushing the student to rely on the video rather than textual cues alone. We then measure the teacher's visual reliance in the same way and compute the \emph{vision-reliance gap}: the difference between the teacher's and student's reliance at each token. A large positive gap identifies positions where the teacher uses visual information more strongly than the student. By using this gap to weight both learning objectives, we focus training on tokens where the visual input is under-utilized. In this way, our method enables the model to ``watch itself watch,'' to identify and improve where its predictions depend on what it sees.

Our main contribution is the novel VIOS post-training paradigm, an efficient framework for video reasoning that combines: a co-evolving EMA teacher that provides an ever-improving privileged target; a contrastive objective that ensures the student policy is visually aware; and a vision reliance gap gate that concentrates learning on the positions where the visual input is under-utilized.

We evaluate VIOS across 12 popular benchmarks spanning general video understanding, temporal reasoning, spatial reasoning, knowledge, and long-video understanding. VIOS consistently outperforms its base models and contemporary video-reasoning systems, while also reaching this performance in over $9\times$ fewer training steps, $60\times$ fewer FLOPs, and $111\times$ fewer generated tokens. Our results suggest that in video reasoning, the path forward is not scaling external supervision or rollouts but teaching models to recognize, on their own, where their reasoning should depend on what they see.

%% file: 2-related-work.tex
\section{Related Work}
\vspace*{-0.1in}
\paragraph{Video Reasoning in Multimodal Large Language Models.}
RL with verifiable rewards has driven recent progress in LLM reasoning~\cite{deepseek_r1, deepseekmath, kimi_k15}, where Group Relative Policy Optimization (GRPO) and related objectives elicit strong multi-step reasoning in math and code. Recent efforts extend these ideas to video reasoning in MLLMs~\cite{video_r1, videochat_r1, videochat_r15, videorft, video_r2, video_ver, arrow_of_time}. Video-R1~\cite{video_r1}, VideoChat-R1~\cite{videochat_r1}, and Video-R2~\cite{video_r2} augment GRPO with spatio-temporal rewards and objectives, while VideoRFT~\cite{videorft} and Video-VER~\cite{video_ver} design rewards that ground reasoning in visual evidence; VideoChat-R1.5~\cite{videochat_r15} instead interleaves iterative perception with GRPO. However, these works are limited by GRPO's sample-inefficiency and sparse reward signal, and some works~\cite{videochat_r1, video_r2, video_ver} also rely on substantially larger or more capable models as judges during GRPO training. Post-training approaches that move beyond GRPO remain underexplored in this setting. VIOS departs from this line of work by obtaining dense, per-token supervision recovered from a single rollout of a single model, which drastically improves sample-efficiency, enabling less computation and faster training time. 

\vspace*{-0.1in}
\paragraph{Visual-Input Sensitivity in Multimodal Models.}
Prior work has explored the sensitivity of MLLM predictions to visual perturbations as a means of guiding decoding and preference optimization. Visual Contrastive Decoding~\cite{leng2024vcd} contrasts next-token logits from an original image and its diffusion-noised counterpart to adjust decoding without training. ConVis~\cite{park2025convis} similarly performs training-free contrastive decoding, but constructs the contrasting image using an external text-to-image model. Token Preference Optimization~\cite{gu2025tpo} instead uses image-corruption sensitivity to weight token-level preferences within a DPO-style objective. However, these methods use visual sensitivity to adjust decoding or weight off-policy preference optimization, rather than directly addressing visual under-utilization during on-policy reasoning. VIOS advances this direction by integrating a visual-awareness contrastive objective with privileged on-policy self-distillation, using the relative teacher-student reliance gap to concentrate dense supervision where the student under-utilizes video evidence.

\vspace*{-0.1in}
\paragraph{On-policy Self-Distillation for Reasoning.}
Knowledge distillation has long been used to transfer a teacher's behavior to a student via soft targets~\cite{hinton_kd, seqkd, distilbert}, but off-policy distillation from static teacher trajectories suffers from distribution shift since the student is never exposed to its own mistakes. On-policy distillation~\cite{agarwal_opd, tml_opd, speculative_kd} mitigates this by sampling trajectories from the student and grading them with dense, token-level teacher supervision, combining the on-policy relevance of RL with a non-sparse learning signal, though it typically requires a separate and more capable teacher model. On-policy self-distillation~\cite{opsd} (OPSD) takes a step toward using a single model as both teacher and student: the student samples on-policy rollouts, while the teacher, conditioned additionally on a privileged ground-truth chain of thought, supplies per-token targets using a fixed-direction KL objective. However, OPSD requires ground-truth reasoning traces, freezes the teacher at the base model, and remains limited to text-only reasoning. VIOS addresses these limitations with a co-evolving EMA teacher conditioned only on the final answer and trained via a symmetric, bounded JSD objective, a visual-awareness contrastive objective that encourages video-dependent predictions, and a per-token vision-reliance gap gate that focuses supervision where vision is under-utilized. Together, these remove the need for ground-truth rationales, lift the rationalization ceiling of a stationary teacher, and explicitly target visual under-utilization in multimodal reasoning.

%% file: 3-method.tex
\section{Method}
\label{method}

\begin{figure}[t]
    \centering
    \includegraphics[width=\textwidth]{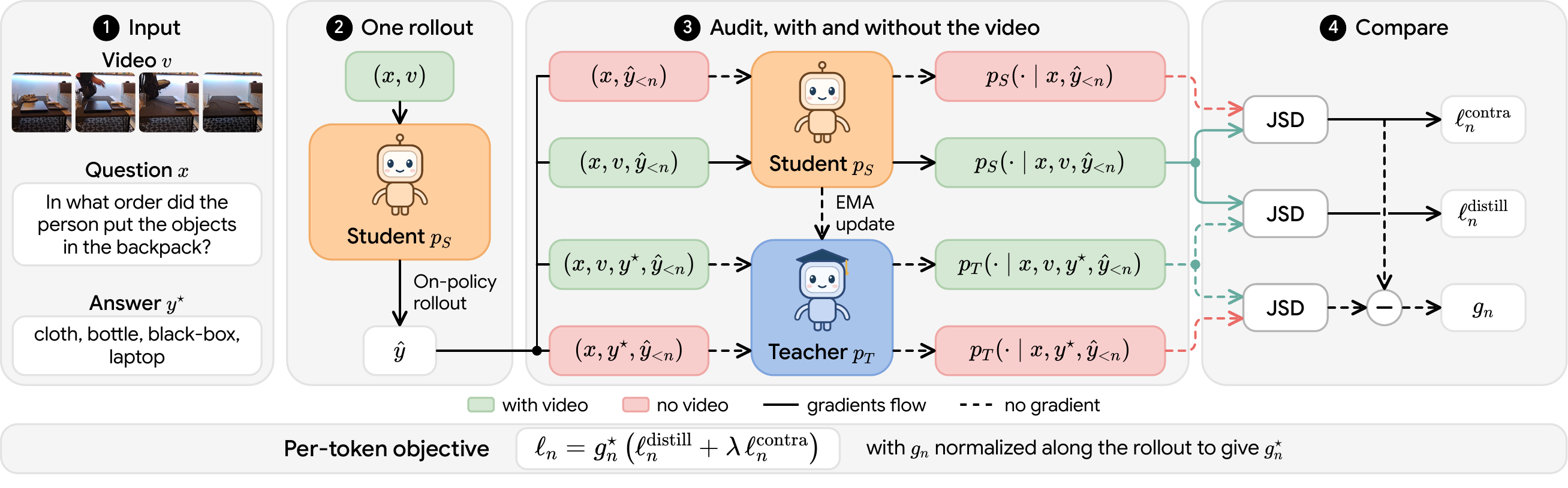}
    \caption{Overview of VIOS. A single on-policy rollout provides per-token, vision-aware supervision through three JSD comparisons between the student $p_S$ and EMA teacher $p_T$: distillation toward the teacher, contrastive separation from the student's no-vision counterpart, and a reliance gap $g_n$ that emphasizes tokens where the teacher uses vision and the student does not.
    }
    \label{fig:method}
\end{figure}
 
A model can audit its own visual reliance: comparing its next-token distribution with and without the vision input at any position measures how much that prediction depends on what was seen. We introduce \textbf{VIOS} (\textbf{Vi}sually-Aware \textbf{O}n-Policy \textbf{S}elf-Distillation), turning this into a post-training procedure. Auditing the evolving student and a privileged teacher exposes a per-token \emph{reliance gap} that reveals where the teacher exploits visual evidence and the student does not. From this we recover dense, vision-aware supervision out of a single on-policy rollout of a single model, with no external reward, judge, or teacher. We instantiate the idea through three components: a self-distillation term against a co-evolving EMA teacher, a contrastive term that forces predictions to depend on $\vv$, and a reliance-gap gate that focuses both on visually-sensitive tokens.
 
\paragraph{Notation.} We consider a video question answering (VQA) dataset $\mathcal{D} = \{(\vx_i, \vv_i, \ystar_i)\}_{i=1}^{N}$, where each example consists of a text query $\vx$, a video $\vv$, and a ground-truth final answer $\ystar$. All policies are instantiated from a single vision-language model with parameters $\theta$.
\begin{itemize}[leftmargin=1.5em,itemsep=1pt,topsep=2pt]
  \item \emph{Student} $\pS(\cdot \mid \vx, \vv)$ and \emph{teacher} $\pT(\cdot \mid \vx, \vv, \ystar)$ share the same architecture. The teacher additionally conditions on the privileged ground-truth answer $\ystar$ and is parameterized by an EMA copy $\bar\theta$ of the student.
  \item $\yhat = (\yhat_1, \ldots, \yhat_{|\yhat|}) \sim \pS(\cdot \mid \vx, \vv)$ denotes an on-policy student rollout, and $p(\cdot \mid \ldots, \yhat_{<n})$ the next-token distribution at position $n$ along it.
  \item $\JSD(p \| q) = \tfrac{1}{2}\KL(p \| m) + \tfrac{1}{2}\KL(q \| m)$ with $m = \tfrac{1}{2}(p+q)$, bounded in $[0, \log 2]$.
  \item $\sg[\cdot]$ denotes stop-gradient.
\end{itemize}
 
\subsection{Self-Distillation Term}

The student samples a rollout $\yhat \sim \pS(\cdot \mid \vx, \vv)$, and the teacher scores every prefix. The teacher sees the same inputs \emph{plus} the ground-truth answer $\ystar$, so its next-token distribution reflects what the trajectory should look like given the correct answer. Aligning the student to the teacher position-by-position along its own rollout converts privileged knowledge of $\ystar$ into a dense, on-policy signal.
 
\paragraph{EMA teacher.} Rather than freezing the teacher at initialization, we set $\bar\theta$ to be an exponential moving average of the student, updated after each optimizer step,
\begin{equation}
  \bar\theta \leftarrow \alpha\, \bar\theta + (1-\alpha)\, \theta, \qquad \alpha \in (0,1).
\end{equation}
A fixed teacher caps the attainable performance at the initial model's rationalization ability. An EMA teacher instead co-evolves with the student, so privileged-context supervision improves as the student's underlying capability improves. The slow averaging preserves the stability of a non-updating teacher (gradients still only flow through the student) while lifting the ceiling imposed by a stationary target.
 
\paragraph{JSD objective.} We measure teacher-student disagreement with the Jensen-Shannon divergence. JSD is symmetric, bounded, and sits between forward KL (mode-covering) and reverse KL (mode-seeking). In video reasoning, many phrasings describe the same content equally well; mode-covering all of them leaves the policy weak on any one, while mode-seeking collapses onto a single rationalized continuation. JSD's balance keeps the student committed to specific phrasings while permitting plausible alternatives. The distillation loss is
\begin{equation}
  \mathcal{L}_{\text{distill}}(\theta) = \E_{(\vx,\vv,\ystar)\sim\mathcal{D}}\; \E_{\yhat \sim \pS(\cdot\mid\vx,\vv)} \!\left[ \frac{1}{|\yhat|} \sum_{n=1}^{|\yhat|} \JSD\!\Big(\pT(\cdot \mid \vx, \vv, \ystar, \yhat_{<n}) \,\big\|\, \pS(\cdot \mid \vx, \vv, \yhat_{<n})\Big) \right].
  \label{eq:distill}
\end{equation}
This is the backbone: at every token, the privileged teacher provides a target and the student is pulled toward it. What it does not yet do is ensure the student's predictions actually depend on $\vv$, nor distinguish positions where vision is decisive from positions where it is irrelevant. The next two subsections supply each piece.
 
\subsection{Visual Awareness Contrastive Term}
\label{sec:vg}

Eq.~\eqref{eq:distill} admits a shortcut solution. The teacher shares the student's architecture and is itself prone to vision under-utilization. With the privileged answer $\ystar$ in context, it is especially well-positioned to produce plausible supervision from text alone, without consulting $\vv$. A student that matches such fabricated supervision can minimize $\mathcal{L}_{\text{distill}}$ with continuations that are textually coherent but under-utilize the video. For example, in Figure \ref{fig:method}'s sample, ``In what order did the person put the objects in the backpack?'', the teacher could confidently encourage the student to generate an ordering independent of the video, instead relying on common packing patterns such as placing a laptop last.
 
We counteract this shortcut with a contrastive term that pushes the student's with-vision distribution \emph{away} from its counterpart that is not conditioned on $\vv$:
\begin{equation}
  \mathcal{L}_{\text{contra}}(\theta) = -\,\E\!\left[ \frac{1}{|\yhat|} \sum_{n=1}^{|\yhat|} \JSD\!\Big(\pS(\cdot \mid \vx, \vv, \yhat_{<n}) \,\big\|\, \sg\!\big[\pS(\cdot \mid \vx, \yhat_{<n})\big]\Big) \right].
  \label{eq:vision}
\end{equation}
Stop-gradient on the vision-ablated branch ensures that we shape the with-vision predictions rather than degrading the no-vision branch. The minus sign expresses that we \emph{maximize} the divergence, so the student's predictions become sensitive to $\vv$. Intuitively, $\mathcal{L}_{\text{distill}}$ specifies \emph{what} the student should predict, while $\mathcal{L}_{\text{contra}}$ specifies that those predictions must be generated by \emph{utilizing} the video.
 
\subsection{Vision Reliance Gap Gate}
\label{sec:vrg}

Finally, our approach distinguishes visually decisive positions. Many token positions do not need supervision: stylistic or purely linguistic choices (``therefore'' vs.\ ``thus''), boilerplate scaffolding, and any stretch of reasoning that follows from $\vx$ and $\yhat_{<n}$ without necessitating $\vv$. Applying divergence pressure on these tokens destabilizes distributions that do not require adjustment, so we prefer to concentrate supervision on positions that actually require the video. As a result, we introduce a gate that directly localizes the signal to visually-sensitive positions.
 
\paragraph{Per-token vision reliance.} At each position $n$, we measure how much each policy's prediction changes when vision is removed from its context:
\begin{align}
  r^{T}_n &= \sg\!\Big[\JSD\!\Big(\pT(\cdot \mid \vx, \vv, \ystar, \yhat_{<n}) \,\big\|\, \pT(\cdot \mid \vx, \ystar, \yhat_{<n})\Big)\Big], \\
  r^{S}_n &= \sg\!\Big[\JSD\!\Big(\pS(\cdot \mid \vx, \vv, \yhat_{<n}) \,\big\|\, \pS(\cdot \mid \vx, \yhat_{<n})\Big)\Big].
\end{align}
Both quantities are detached from the computation graph. The proposed \emph{vision reliance gap} at token $n$ is
\begin{equation}
  g_n = r^{T}_n - r^{S}_n.
\end{equation}
A large positive $g_n$ means the teacher's prediction at this position genuinely depends on the video while the student's does not. These are precisely the tokens where the student is under-utilizing the vision input and where our objective should apply the strongest pressure. Tokens for which $g_n \approx 0$ are ones where teacher and student agree on how much vision matters, whether because the position is stylistic, text-only derivable, or the student is correctly using the video, and are less emphasized.
 
\paragraph{Two-stage normalization.} Raw JSD values over a large vocabulary are numerically small and concentrate near zero, so multiplying the per-token losses by $g_n$ directly would prevent propagation of the training signal. We therefore apply a two-stage normalization along each trajectory. First, min-max rescaling stretches $g_n$ to the unit interval,
\begin{equation}
  \tilde{g}_n = \frac{g_n - \min_k g_k}{\max_k g_k - \min_k g_k},
\end{equation}
which recovers the relative structure of visually-sensitive positions independent of the absolute JSD magnitude. Second, we rescale so the trajectory mean of the gate is one,
\begin{equation}
  g^{\star}_n = \frac{\tilde{g}_n}{\tfrac{1}{|\yhat|}\sum_{k=1}^{|\yhat|} \tilde{g}_k}.
  \label{eq:gate}
\end{equation}
This keeps the per-trajectory loss magnitude in the same regime as the ungated baseline and prevents the gate from affecting the learning rate.
 
\paragraph{Gated objective.} The gate is applied element-wise to both loss terms. Writing $\ell^{\text{distill}}_n$ and $\ell^{\text{contra}}_n$ for the per-position integrands of Eqs.~\eqref{eq:distill} and \eqref{eq:vision}, the full objective is
\begin{equation}
  \mathcal{L}(\theta) = \E\!\left[\frac{1}{|\yhat|} \sum_{n=1}^{|\yhat|} g^{\star}_n \!\left( \ell^{\text{distill}}_n + \lambda\, \ell^{\text{contra}}_n \right)\right],
  \label{eq:final}
\end{equation}
with $\lambda > 0$ controlling the vision-aware strength and $g^{\star}_n$ treated as a constant during backpropagation. The gate increases the distillation and awareness signals on the tokens where the teacher exploits visual input and the student does not.

%% file: 4-experiments.tex
\section{Experiments}

We evaluate VIOS on 12 video understanding benchmarks against contemporary video reasoning systems and across 3 base models, spanning model families and sizes, then examine training efficiency and ablate each component of the method. Section~\ref{sec:setup} describes the setup. Section~\ref{sec:main} reports the main comparison. Section~\ref{sec:efficiency} examines training efficiency. Section~\ref{sec:ablation} ablates each component independently.

\subsection{Setup}
\label{sec:setup}
 
\paragraph{Training Strategy.}
We apply VIOS to Qwen2.5-VL-7B~\cite{qwen25vl}, InternVL2.5-8B, and InternVL2.5-4B~\cite{internvl25} using the Video-R1-260k corpus~\cite{video_r1}. Because VIOS distills along on-policy student rollouts, the base policy must be able to produce a chain of thought. Therefore, for Qwen2.5-VL-7B we initialize from Qwen2.5-VL-7B-SFT, the SFT-tuned base model released with Video-R1~\cite{video_r1}.
 
\paragraph{Benchmarks.}
We evaluate on 12 public video understanding benchmarks spanning five categories, chosen to cover the full breadth of capabilities required of a modern video MLLM.\footnote{We also evaluate on four static image benchmarks covering perception, charts, and knowledge~\cite{blink, chartqa, cvbench, mmmu} (Appendix~\ref{app:image_benchmarks}).} For \textit{general video understanding} we use MVBench~\cite{mvbench}, Video-MME~\cite{video-mme}, and MINERVA~\cite{minerva}. For \textit{knowledge-intensive reasoning} we use MMVU~\cite{mmvu}, Video-MMMU~\cite{video-mmmu}, VideoMathQA~\cite{videomathqa}, and SciVideoBench~\cite{scivideobench}. For \textit{temporal reasoning} we use TempCompass~\cite{tempcompass} and TVBench~\cite{tvbench}. For \textit{long video understanding} we use LongVideoBench~\cite{longvideobench} and MLVU~\cite{mlvu}. For \textit{spatial reasoning} we use VSI-Bench~\cite{vsibench}. Together these benchmarks probe short and long horizons, abstract knowledge and concrete perception, and a wide range of question formats. They form a broad and widely studied cross section of video understanding, so consistent gains across them are intended to validate VIOS as a fundamental improvement to post-training for modern video models. Following Video-R1, we use the multiple-choice subset of MMVU. Subtitles are excluded from Video-MME.
 
\paragraph{Implementation.}
During training we cap the number of video frames at 16 and the number of vision tokens per frame at 128. At inference these are doubled to 32 frames and 256 tokens per frame with temperature set to 0.01 and top\_p set to 0.001. The contrastive term in Eq.~\eqref{eq:final} uses $\lambda = 0.5$. The EMA teacher uses decay $\alpha = 0.99$. Gate normalization and stop-gradient placement follow Section~\ref{method}. Each model is trained for 100 VIOS steps on 8 NVIDIA H200 GPUs. See Appendix~\ref{app:hyperparams} for hyperparameters, Appendix~\ref{app:lambda_alpha_ablations} for $\lambda$ and $\alpha$ ablations, and Appendix~\ref{app:eval_protocol} for evaluation.
 
\paragraph{Baselines.}
We compare VIOS against the corresponding base model across Qwen2.5-VL-7B-SFT~\cite{qwen25vl,video_r1}, InternVL2.5-8B, and InternVL2.5-4B~\cite{internvl25} to isolate the contribution of VIOS post-training. Qwen2.5-VL-7B is the backbone most widely adopted by contemporary video reasoning work, so anchoring our study to the same starting point ensures a fair comparison against recent work, while varying the backbone additionally shows the generality of our idea. We compare against contemporary video reasoning models that span the design space of recent work. Video-R1~\cite{video_r1}, VideoChat-R1~\cite{videochat_r1}, and Video-R2~\cite{video_r2} augment GRPO with spatio-temporal rewards and objectives. VideoRFT~\cite{videorft} designs rewards that ground reasoning in visual evidence. VideoChat-R1.5~\cite{videochat_r15} instead interleaves iterative perception with GRPO. Together these methods cover the dominant approaches to post-training video reasoners and provide a representative point of reference for evaluating VIOS.

\subsection{Main Results}
\label{sec:main}

\definecolor{rowgray}{gray}{0.92}
\newcolumntype{B}{>{\centering\arraybackslash}p{0.75cm}}
\newcommand{\rh}[1]{\makebox[0pt][c]{\rotatebox{90}{#1\strut}}}
 
\begin{table}[t]
\centering
\scriptsize
\setlength{\tabcolsep}{1.5pt}
\renewcommand{\arraystretch}{1.15}
\begin{tabular}{l BBB BBBB BB BB B B}
\toprule
& \multicolumn{3}{c}{\makebox[0pt][c]{\textbf{General}}}
  & \multicolumn{4}{c}{\makebox[0pt][c]{\textbf{Knowledge}}}
  & \multicolumn{2}{c}{\makebox[0pt][c]{\textbf{Temporal}}}
  & \multicolumn{2}{c}{\makebox[0pt][c]{\textbf{Long Video}}}
  & \makebox[0pt][c]{\textbf{Spatial}}
  & \\
\cmidrule(lr){2-4} \cmidrule(lr){5-8} \cmidrule(lr){9-10} \cmidrule(lr){11-12} \cmidrule(lr){13-13}
\textbf{Model}
& \rh{MVBench} & \rh{Video-MME} & \rh{MINERVA}
& \rh{MMVU} & \rh{Video-MMMU} & \rh{VideoMathQA} & \rh{SciVideoBench}
& \rh{TempCompass} & \rh{TVBench}
& \rh{LongVideoBench} & \rh{MLVU}
& \rh{VSI-Bench}
& \rh{Average} \\
\midrule
Video-R1~\cite{video_r1}        & \underline{64.42} & \underline{58.37} & \underline{30.76} & 63.04 & \textbf{49.22} & 23.57 & 24.50 & 71.58 & 50.81 & \underline{55.42} & \textbf{63.11} & \underline{35.25} & \underline{49.17} \\
VideoRFT~\cite{videorft}        & 62.25 & 58.26 & 28.26 & \underline{64.96} & 47.44 & 24.76 & 24.10 & \underline{72.65} & \textbf{53.35} & 54.08 & 61.82 & 35.10 & 48.92 \\
VideoChat-R1~\cite{videochat_r1}    & 62.40 & 57.07 & 29.47 & 63.68 & 47.67 & 23.33 & 23.30 & 72.43 & 52.16 & 53.85 & 57.82 & 33.08 & 48.02 \\
VideoChat-R1.5~\cite{videochat_r15}  & 62.17 & 56.89 & 28.56 & 63.20 & \textbf{49.22} & \underline{25.00} & 23.20 & 72.28 & 48.40 & 54.82 & 58.05 & 31.95 & 47.81 \\
Video-R2~\cite{video_r2}        & 62.08 & 56.04 & 27.42 & 64.80 & \underline{48.33} & 20.00 & \textbf{25.40} & 72.64 & 51.68 & 50.19 & 58.00 & 32.56 & 47.43 \\
Qwen2.5-VL-SFT~\cite{qwen25vl, video_r1}  & 60.72 & 55.15 & 29.55 & 59.36 & 47.22 & 23.10 & 22.40 & 67.25 & 49.82 & 52.51 & 55.70 & 33.44 & 46.35 \\
\rowcolor{rowgray}
+ VIOS (Ours)   & \textbf{65.50} & \textbf{59.59} & \textbf{31.36} & \textbf{65.76} & 47.44 & \textbf{25.48} & \underline{24.90} & \textbf{72.77} & \underline{52.24} & \textbf{55.87} & \underline{62.74} & \textbf{35.73} & \textbf{49.95} \\

\midrule
InternVL2.5-4B~\cite{internvl25}
& 61.27 & 53.19 & 27.12 & 52.96 & 34.89 & 19.52 & 15.90 & 62.25 & 52.40 & 43.68 & 57.64 & 33.72 & 42.88 \\
\rowcolor{rowgray}
+ VIOS (Ours)
& \textbf{68.42} & \textbf{57.15} & \textbf{30.91} & \textbf{57.44} & \textbf{40.44} & \textbf{21.19} & \textbf{19.80} & \textbf{68.01} & \textbf{57.43} & \textbf{51.61} & \textbf{63.57} & \textbf{34.14} & \textbf{47.51} \\

\midrule
InternVL2.5-8B~\cite{internvl25}
& 62.42 & 57.93 & 30.38 & 56.48 & 43.56 & 19.29 & 17.90 & 70.12 & 57.11 & 51.16 & 61.96 & \textbf{38.04} & 47.20 \\
\rowcolor{rowgray}
+ VIOS (Ours)
& \textbf{69.95} & \textbf{61.52} & \textbf{33.18} & \textbf{60.96} & \textbf{45.89} & \textbf{23.81} & \textbf{22.50} & \textbf{72.21} & \textbf{61.74} & \textbf{58.71} & \textbf{66.79} & 36.33 & \textbf{51.13} \\

\bottomrule
\end{tabular}
\caption{Main results across 12 video understanding benchmarks. VIOS consistently improves all three backbone models, achieving average gains of 3.60, 3.93, and 4.63 points. On Qwen2.5-VL~\cite{qwen25vl, video_r1}, VIOS ranks first or second on 11 of 12 benchmarks while requiring over $9\times$ fewer training steps than the baselines (e.g. Video-R1~\cite{video_r1} requires 1,000 training steps).}
\label{tab:video_benchmarks}
\end{table}

Table~\ref{tab:video_benchmarks} reports performance across the 12 benchmarks. On Qwen2.5-VL~\cite{qwen25vl, video_r1}, VIOS improves the average by 3.60 points, ranking first on 8 benchmarks (MVBench~\cite{mvbench}, Video-MME~\cite{video-mme}, MINERVA~\cite{minerva}, MMVU~\cite{mmvu}, VideoMathQA~\cite{videomathqa}, TempCompass~\cite{tempcompass}, LongVideoBench~\cite{longvideobench}, and VSI-Bench~\cite{vsibench}) and second on 3 more (TVBench~\cite{tvbench}, MLVU~\cite{mlvu}, and SciVideoBench~\cite{scivideobench}). The largest gains over the base model are on MLVU (+7.0), MMVU (+6.4), TempCompass (+5.5), MVBench (+4.8), and Video-MME (+4.4). These improvements generalize across model families, with VIOS boosting the average performance of InternVL2.5-4B and InternVL2.5-8B~\cite{internvl25} by 4.63 and 3.93 points, respectively. Notably, VIOS achieves these results using a single on-policy rollout per example, no external judge model, and over $9\times$ fewer training steps than GRPO-based systems (e.g. Video-R1~\cite{video_r1}, which requires 1,000 training steps).

\begin{figure}[t]
    \centering
    \includegraphics[width=\textwidth]{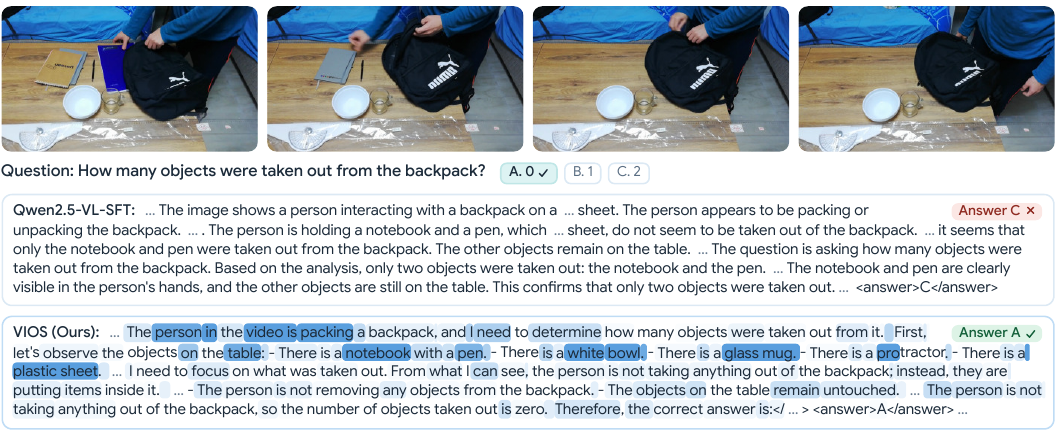}
    \includegraphics[width=\textwidth]{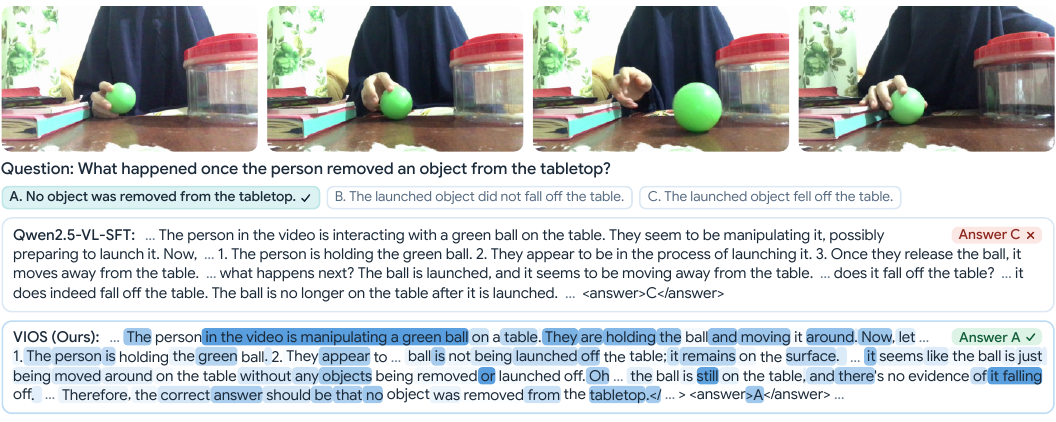}
    \caption{Comparison of reasoning traces between base model and VIOS. The VIOS rollout is highlighted representing the per-token loss at that position, with darker indicating higher loss.}
    \label{fig:qualitative_examples}
\end{figure}

Figure \ref{fig:qualitative_examples} presents qualitative examples of the thinking chains generated by the base model, Qwen2.5-VL-SFT~\cite{qwen25vl, video_r1}, and its VIOS post-trained counterpart. The loss correctly captures the tokens that are relevant to the video and to answering the questions. In the first example, the loss highlights each of the items on the table and notes that the individual is packing, and in the second example the loss highlights the individual's action upon the ball and identifies where the student mentioned no evidence of the ball falling off. It is important to note that high loss may mean the teacher assigned a higher probability to that token than the student, not that VIOS necessarily thinks those tokens are incorrect. 

\subsection{Efficiency Comparison}
\label{sec:efficiency}

We next quantify VIOS's training cost relative to GRPO-based baselines, specifically vanilla GRPO and Video-R1~\cite{video_r1} since we utilize the same training corpus. For a fair comparison, all experiments in this section are initialized from Qwen2.5-VL-SFT~\cite{qwen25vl, video_r1}, the base model used by Video-R1. We rerun VIOS with the same per-step training configuration used by Video-R1 (4 GPUs, per-device batch size 1, learning rate $1\!\times\!10^{-6}$). Since this reduces the samples processed per step, we run 110 iterations rather than 100. Both baselines are obtained identically following Video-R1's publicly available training scripts. For vanilla GRPO, we disable all of Video-R1's GRPO augmentations. Even with the slight increase in steps, VIOS still trains for over $9\times$ fewer steps than Video-R1.

\begin{figure}[t]
    \centering
    \includegraphics[width=\textwidth]{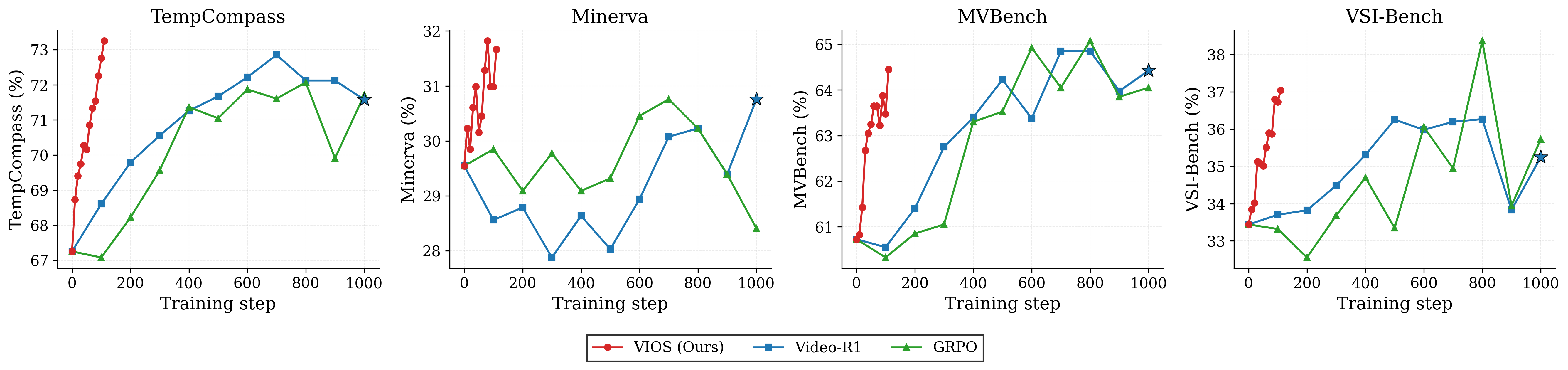}
    \caption{Accuracy across training on four representative benchmarks. VIOS matches or exceeds Video-R1's final accuracy within roughly 100 steps, compared with approximately 1000 steps for Video-R1 and GRPO. Stars indicate the performance of the public Video-R1 checkpoint.}
    \label{fig:training_vs_step}
\end{figure}
\begin{figure}[t]
    \centering
    \includegraphics[width=\textwidth]{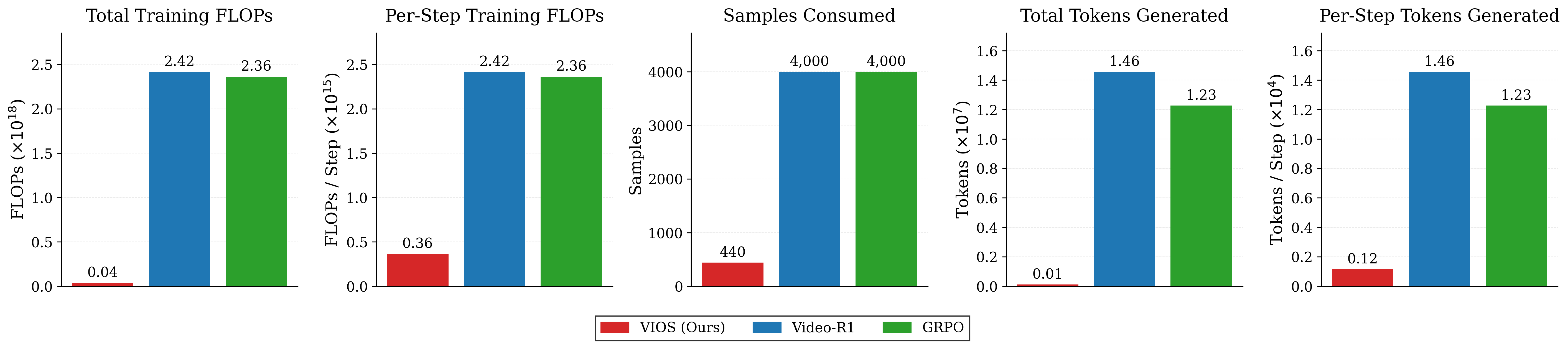}
    \caption{Training cost comparison of VIOS, Video-R1, and GRPO. VIOS substantially reduces total and per-step compute, sample consumption, and token generation through a single on-policy rollout per step. See Appendix~\ref{app:flop-accounting} for FLOP calculations.}
    \label{fig:bar_figs}
\end{figure}
 
\begin{table}[!t]
\centering
\scriptsize
\setlength{\tabcolsep}{1.5pt}
\renewcommand{\arraystretch}{1.15}
\begin{tabular}{l c BBB BBBB BB BB B B}
\toprule
& & \multicolumn{3}{c}{\makebox[0pt][c]{\textbf{General}}}
  & \multicolumn{4}{c}{\makebox[0pt][c]{\textbf{Knowledge}}}
  & \multicolumn{2}{c}{\makebox[0pt][c]{\textbf{Temporal}}}
  & \multicolumn{2}{c}{\makebox[0pt][c]{\textbf{Long Video}}}
  & \makebox[0pt][c]{\textbf{Spatial}}
  & \\
\cmidrule(lr){3-5} \cmidrule(lr){6-9} \cmidrule(lr){10-11} \cmidrule(lr){12-13} \cmidrule(lr){14-14}
\textbf{Model} & \textbf{Steps}
& \rh{MVBench} & \rh{Video-MME} & \rh{MINERVA}
& \rh{MMVU} & \rh{Video-MMMU} & \rh{VideoMathQA} & \rh{SciVideoBench}
& \rh{TempCompass} & \rh{TVBench}
& \rh{LongVideoBench} & \rh{MLVU}
& \rh{VSI-Bench}
& \rh{Average} \\
\midrule
Video-R1                                  & 1000 & 64.42 & 58.37 & 30.76 & 63.04 & \textbf{49.22} & 23.57 & 24.50 & 71.58 & 50.81 & \textbf{55.42} & \textbf{63.11} & 35.25 & 49.17 \\
\rowcolor{rowgray}
VIOS\textsuperscript{$\dagger$} (Ours)    & \textbf{110} & \textbf{64.45} & \textbf{58.63} & \textbf{31.67} & \textbf{65.12} & 46.33 & \textbf{23.60} & \textbf{25.24} & \textbf{73.25} & \textbf{52.00} & 54.00 & 61.68 & \textbf{37.05} & \textbf{49.42} \\
\bottomrule
\end{tabular}
\caption{Under Video-R1's training configuration, VIOS achieves higher average accuracy and outperforms Video-R1 on 9 of 12 benchmarks in over $9\times$ fewer steps.}
\label{tab:efficiency}
\end{table}

Table~\ref{tab:efficiency} reports results under this configuration. VIOS matches or exceeds Video-R1~\cite{video_r1} on 9 out of 12 benchmarks and outperforms on average. Figure~\ref{fig:training_vs_step} traces accuracy as a function of training step on four representative benchmarks for VIOS, Video-R1, and a GRPO baseline trained from the same starting point. VIOS gains performance at a consistent and very accelerated pace compared to the two baselines that require close to 1000 steps to reach comparable accuracy. Figure~\ref{fig:bar_figs} reports five complementary measures of training cost. VIOS consumes a small fraction of the total compute, samples, and generation of either baseline. Specifically, it uses 60$\times$ fewer total training FLOPs, 6.7$\times$ fewer per-step training FLOPs, 111$\times$ fewer total tokens generated, and 12$\times$ fewer per-step tokens generated than Video-R1~\cite{video_r1}.

\subsection{Ablation Study}
\label{sec:ablation}
 
We ablate the three components of VIOS independently: the EMA teacher, the visual awareness contrastive objective, and the visual reliance gap gate. All variants stem from the main Qwen2.5-VL-SFT~\cite{qwen25vl, video_r1} configuration detailed in Section~\ref{sec:setup}.
 
\paragraph{Teacher update rule.}
Table~\ref{tab:ablation_teacher} compares an EMA-updated teacher with a frozen teacher held at the base model. Both improve substantially over SFT, but the EMA teacher consistently outperforms the frozen one, by 0.7 points on average and on nine of twelve benchmarks. The largest gains occur on reasoning-intensive benchmarks (VideoMathQA, Video-MME, MINERVA, MLVU), where the frozen teacher's rationalization ability imposes a fixed ceiling. By co-evolving with the student, the EMA teacher avoids this ceiling in principle. The empirical gap can be attributed to our training budget of 100 iterations with EMA decay $\alpha = 0.99$ which limits the compounding of co-evolution.
 
\begin{table}[t]
\centering
\scriptsize
\setlength{\tabcolsep}{1.5pt}
\renewcommand{\arraystretch}{1.15}
\begin{tabular}{l BBB BBBB BB BB B B}
\toprule
& \multicolumn{3}{c}{\makebox[0pt][c]{\textbf{General}}}
  & \multicolumn{4}{c}{\makebox[0pt][c]{\textbf{Knowledge}}}
  & \multicolumn{2}{c}{\makebox[0pt][c]{\textbf{Temporal}}}
  & \multicolumn{2}{c}{\makebox[0pt][c]{\textbf{Long Video}}}
  & \textbf{Spatial}
  & \\
\cmidrule(lr){2-4} \cmidrule(lr){5-8} \cmidrule(lr){9-10} \cmidrule(lr){11-12} \cmidrule(lr){13-13}
\textbf{Variant}
& \rh{MVBench} & \rh{Video-MME} & \rh{MINERVA}
& \rh{MMVU} & \rh{Video-MMMU} & \rh{VideoMathQA} & \rh{SciVideoBench}
& \rh{TempCompass} & \rh{TVBench}
& \rh{LongVideoBench} & \rh{MLVU}
& \rh{VSI-Bench}
& \rh{Average} \\
\midrule
Qwen2.5-VL-SFT             & 60.72 & 55.15 & 29.55 & 59.36 & 47.22 & 23.10 & 22.40 & 67.25 & 49.82 & 52.51 & 55.70 & 33.44 & 46.35 \\
Frozen Teacher     & \underline{65.07} & \underline{58.15} & \underline{29.92} & \underline{65.60} & \textbf{47.78} & \underline{23.81} & \underline{24.10} & \textbf{72.92} & \textbf{52.71} & \underline{54.90} & \underline{61.50} & \underline{34.90} & \underline{49.28} \\
\rowcolor{rowgray}
EMA Teacher (Ours) & \textbf{65.50} & \textbf{59.59} & \textbf{31.36} & \textbf{65.76} & \underline{47.44} & \textbf{25.48} & \textbf{24.90} & \underline{72.77} & \underline{52.24} & \textbf{55.87} & \textbf{62.74} & \textbf{35.73} & \textbf{49.95} \\
\bottomrule
\end{tabular}
\caption{Ablation on the teacher update rule. The EMA teacher outperforms the frozen teacher on 9 of 12 benchmarks, with the largest gaps on reasoning-heavy benchmarks.}
\label{tab:ablation_teacher}
\end{table}
 
\paragraph{Visual reliance gap gate.}
Table~\ref{tab:ablation_gate} compares our visual reliance gap gate against an ungated distillation loss. The gated configuration wins on 9 of 12 benchmarks and improves the average by 0.6 points, with the largest gains on VSI-Bench~\cite{vsibench} (+2.6), MLVU~\cite{mlvu} (+1.8), and Video-MME~\cite{video-mme} (+1.5). These results demonstrate the benefit of selectively weighting supervision rather than treating all token positions equally. By emphasizing positions where the teacher and student differ in visual reliance, the gate directs learning toward predictions where the student may be under-utilizing vision.
 
\begin{table}[t]
\centering
\scriptsize
\setlength{\tabcolsep}{1.5pt}
\renewcommand{\arraystretch}{1.15}
\begin{tabular}{l BBB BBBB BB BB B B}
\toprule
& \multicolumn{3}{c}{\makebox[0pt][c]{\textbf{General}}}
  & \multicolumn{4}{c}{\makebox[0pt][c]{\textbf{Knowledge}}}
  & \multicolumn{2}{c}{\makebox[0pt][c]{\textbf{Temporal}}}
  & \multicolumn{2}{c}{\makebox[0pt][c]{\textbf{Long Video}}}
  & \textbf{Spatial}
  & \\
\cmidrule(lr){2-4} \cmidrule(lr){5-8} \cmidrule(lr){9-10} \cmidrule(lr){11-12} \cmidrule(lr){13-13}
\textbf{Variant}
& \rh{MVBench} & \rh{Video-MME} & \rh{MINERVA}
& \rh{MMVU} & \rh{Video-MMMU} & \rh{VideoMathQA} & \rh{SciVideoBench}
& \rh{TempCompass} & \rh{TVBench}
& \rh{LongVideoBench} & \rh{MLVU}
& \rh{VSI-Bench}
& \rh{Average} \\
\midrule
Qwen2.5-VL-SFT       & 60.72 & 55.15 & 29.55 & 59.36 & \underline{47.22} & 23.10 & 22.40 & 67.25 & 49.82 & 52.51 & 55.70 & \underline{33.44} & 46.35 \\
No Gate              & \underline{64.60} & \underline{58.11} & \underline{30.45} & \underline{64.96} & 47.00 & \textbf{27.86} & \textbf{25.10} & \underline{72.19} & \textbf{52.28} & \underline{55.80} & \underline{60.95} & 33.13 & \underline{49.37} \\
\rowcolor{rowgray}
Gate (Ours)          & \textbf{65.50} & \textbf{59.59} & \textbf{31.36} & \textbf{65.76} & \textbf{47.44} & \underline{25.48} & \underline{24.90} & \textbf{72.77} & \underline{52.24} & \textbf{55.87} & \textbf{62.74} & \textbf{35.73} & \textbf{49.95} \\
\bottomrule
\end{tabular}
\caption{Ablation on the selection gate. The reliance gap gate outperforms ungated distillation on 9 of 12 benchmarks.}
\label{tab:ablation_gate}
\end{table}
 
\paragraph{Training-objective components.}
Table~\ref{tab:ablation_loss} adds the visual reliance gap gate and the visual awareness contrastive objective cumulatively over distillation. Each component contributes a positive increment that compounds. Distillation alone improves the base model by 1.9 points by giving the student dense per-token targets. The gate adds another 0.5 points by focusing those targets on visually-sensitive positions. The contrastive term then adds a further 1.3 points by penalizing vision under-utilization to reach full VIOS performance. Its gains span benchmark categories (MVBench~\cite{mvbench} +2.2, VideoMathQA~\cite{videomathqa} +2.4, Video-MME~\cite{video-mme} +1.9, MMVU~\cite{mmvu} +1.8, MLVU~\cite{mlvu} +1.7), reflecting that a privileged-answer teacher can produce plausible supervision from text alone across many task types and that forcing the student's distribution to depend on the video closes that loophole.
 
\newcolumntype{D}{>{\centering\arraybackslash}p{0.70cm}}
\begin{table}[!t]
\centering
\scriptsize
\setlength{\tabcolsep}{1.5pt}
\renewcommand{\arraystretch}{1.15}
\begin{tabular}{l DDD DDDD DD DD D D}
\toprule
& \multicolumn{3}{c}{\makebox[0pt][c]{\textbf{General}}}
  & \multicolumn{4}{c}{\makebox[0pt][c]{\textbf{Knowledge}}}
  & \multicolumn{2}{c}{\makebox[0pt][c]{\textbf{Temporal}}}
  & \multicolumn{2}{c}{\makebox[0pt][c]{\textbf{Long Video}}}
  & \textbf{Spatial}
  & \\
\cmidrule(lr){2-4} \cmidrule(lr){5-8} \cmidrule(lr){9-10} \cmidrule(lr){11-12} \cmidrule(lr){13-13}
\textbf{Variant}
& \rh{MVBench} & \rh{Video-MME} & \rh{MINERVA}
& \rh{MMVU} & \rh{Video-MMMU} & \rh{VideoMathQA} & \rh{SciVideoBench}
& \rh{TempCompass} & \rh{TVBench}
& \rh{LongVideoBench} & \rh{MLVU}
& \rh{VSI-Bench}
& \rh{Average} \\
\midrule
Qwen2.5-VL-SFT                       & 60.72 & 55.15 & 29.55 & 59.36 & 47.22 & 23.10 & 22.40 & 67.25 & 49.82 & 52.51 & 55.70 & 33.44 & 46.35 \\
Distill.                             & \underline{64.23} & 56.96 & \underline{31.06} & 62.56 & \textbf{48.00} & \underline{24.29} & 24.50 & 69.88 & 49.98 & 53.25 & 59.57 & 34.23 & 48.21 \\
Distill. + Gate                      & 63.32 & \underline{57.74} & 30.15 & \underline{64.00} & 46.11 & 23.10 & \underline{24.70} & \underline{70.93} & \textbf{52.55} & \underline{54.23} & \underline{61.09} & \textbf{36.41} & \underline{48.69} \\
\rowcolor{rowgray}
Distill. + Gate + Contra. (Ours)     & \textbf{65.50} & \textbf{59.59} & \textbf{31.36} & \textbf{65.76} & \underline{47.44} & \textbf{25.48} & \textbf{24.90} & \textbf{72.77} & \underline{52.24} & \textbf{55.87} & \textbf{62.74} & \underline{35.73} & \textbf{49.95} \\
\bottomrule
\end{tabular}
\caption{Ablation on training-objective components. Adding distillation, gating, and contrastive learning progressively improves average accuracy, demonstrating the importance of each component.}
\label{tab:ablation_loss}
\end{table}

%% file: 5-conclusion.tex
\section{Conclusion}

We presented VIOS, a post-training procedure that recovers dense, per-token, vision-aware supervision from a single on-policy rollout of a model, with no external reward, judge, or larger teacher. The method follows from a single observation: a model can audit its own visual reliance by comparing its next-token distribution with and without the video in context, and the gap between this quantity for a privileged teacher and a student rollout localizes the positions where the student under-utilizes what it saw. Across 12 video understanding benchmarks, VIOS consistently improves three different backbones, while requiring over $9\times$ fewer training steps, $60\times$ fewer training FLOPs, and $111\times$ fewer tokens generated than contemporary video reasoning systems. These results suggest that the supervision needed for ensuring vision utilization is available in the model itself, recoverable by having it watch itself watch. Extending this principle of self-auditing input reliance beyond video, to other modalities and to multimodal agents, is a promising direction for future work.

\section{Limitations and Future Work}
\label{sec:limitations}

\paragraph{Scale and initialization.}
Our experiments are conducted on Qwen2.5-VL-7B-SFT~\cite{qwen25vl, video_r1}, InternVL2.5-4B, and InternVL2.5-8B~\cite{internvl25} and the Video-R1-260k~\cite{video_r1} training corpus. While this matches the setup used by recent video reasoning systems and isolates the contribution of post-training, it leaves open how VIOS scales to substantially larger models. VIOS also requires a base model that can already produce a coherent chain of thought, which is why for Qwen2.5-VL we initialize from an SFT checkpoint. Testing larger models and less reasoning-ready initializations is a natural next step to assess whether the per-step compute savings continue to compound and how broadly VIOS applies.

\paragraph{Training cost and supervision.}
Computing the reliance gap requires evaluating the student and teacher both with and without the video at every position, roughly doubling the per-step forward cost relative to standard distillation, although this overhead is more than offset by the elimination of multi-rollout RL (Section~\ref{sec:efficiency}). Our setting also assumes a privileged ground-truth final answer is available for each training example, which restricts VIOS to VQA-style data with ground-truth answers. Future work could make reliance-gap estimation more efficient and relax this answer requirement to extend VIOS to open-ended video understanding.

\paragraph{Grounding and correctness.}
On the objective itself, the contrastive term promotes visual utilization but does not enforce grounding to specific frames or regions, so hallucinations consistent with the video's general content can still occur. Correctness enters only implicitly through the privileged-answer teacher rather than as explicit supervision on the final answer, so VIOS inherits whatever rationalization biases the teacher exhibits at any given training step. Addressing these limitations is a promising direction for future work, whether by adding an explicit grounding objective tied to specific frames or regions or incorporating a verifiable correctness signal alongside the privileged-answer teacher.

\paragraph{Broader extensions and diagnostics.}
More broadly, the self-auditing principle behind VIOS---measuring how a prediction shifts when a particular input is removed---is not specific to video. Applying it to other modalities such as audio, depth, and structured sensor streams, as well as to multimodal agents whose decisions must depend on what they perceive, is a direct extension. The reliance gap could also serve as a diagnostic in its own right, providing a token-level audit of whether existing video reasoners actually use their visual input and complementing coarse benchmark accuracy. Exploring these extensions could help establish how broadly self-auditing input reliance generalizes beyond video reasoning.

\section*{Acknowledgements}

This research has been supported by computing support on the Vista GPU Cluster through the Center for Generative AI (CGAI) and the Texas Advanced Computing Center (TACC) at the University of Texas at Austin. UT Austin is supported in part by the IFML NSF AI Institute and Lyda Hill Philanthropies. We thank the UT Vision group for their valuable feedback and insightful discussions. The authors declare no competing interests.

%% file: appendix.tex
\section{Training FLOP Accounting}
\label{app:flop-accounting}

We estimate the per-step FLOPs in Figure~\ref{fig:bar_figs} using the standard $2N/6N$ per-token approximation~\cite{hoffmann2022chinchilla, kaplan2020scaling}: $2N$ for forward-only passes and $6N$ for forward-and-backward passes, where $N$ excludes embedding parameters. Since MLLMs route language and vision tokens through separate modules, we count their FLOPs independently, using $N_{\mathrm{LLM}}=6{,}525{,}621{,}760$ and $N_{\mathrm{ViT}}=675{,}044{,}864$ for Qwen2.5-VL-7B~\cite{qwen25vl}. Let $\tau^p_{\mathrm{L}}$ and $\tau^p_{\mathrm{V}}$ denote the total language and vision tokens processed in pass $p$, summed over all relevant trajectories. The per-step cost is
\begin{equation*}
\begin{aligned}
\mathrm{FLOPs}={}&
\underbrace{2\bigl(N_{\mathrm{LLM}}\tau^{\mathrm{pre}}_{\mathrm{L}}+N_{\mathrm{ViT}}\tau^{\mathrm{pre}}_{\mathrm{V}}\bigr)}_{\text{rollout prefill}}
+\underbrace{2N_{\mathrm{LLM}}\tau^{\mathrm{dec}}_{\mathrm{L}}}_{\text{rollout decode}}\\
&+\underbrace{6\bigl(N_{\mathrm{LLM}}\tau^{\mathrm{pol}}_{\mathrm{L}}+N_{\mathrm{ViT}}\tau^{\mathrm{pol}}_{\mathrm{V}}\bigr)}_{\text{policy update}}
+\underbrace{2\bigl(N_{\mathrm{LLM}}\tau^{\mathrm{aux}}_{\mathrm{L}}+N_{\mathrm{ViT}}\tau^{\mathrm{aux}}_{\mathrm{V}}\bigr)}_{\text{additional forwards}}.
\end{aligned}
\end{equation*}
Here, $\tau^{\mathrm{aux}}=\tau^{\mathrm{tch}}+\tau^{\mathrm{con}}$ for VIOS and $\tau^{\mathrm{aux}}=\tau^{\mathrm{ref}}$ for the GRPO~\cite{deepseekmath} baselines (for both token types). Autoregressive decoding is language-only because the visual prefix is cached.

\begin{table}[h]
\centering
\scriptsize
\setlength{\tabcolsep}{9pt}
\renewcommand{\arraystretch}{1.15}
\begin{tabular}{l p{0.20\linewidth} p{0.20\linewidth} p{0.26\linewidth}}
\toprule
\textbf{Pass} & \textbf{VIOS} & \textbf{Vanilla GRPO} & \textbf{Video-R1} \\
\midrule
Rollout prefill
& Prompt and video
& Shared prompt and video
& Shared policy prefill; 4 temporal-reward prefills \\
Rollout decode
& 1 trajectory
& 8 trajectories
& 8 policy + 4 temporal-reward trajectories \\
Policy update
& 1 trajectory
& 8 trajectories
& 8 policy trajectories \\
Additional forwards
& EMA teacher with/without video; student no-video contrastive pass
& Frozen reference model for KL on 8 trajectories
& Frozen reference model for KL on 8 policy trajectories \\
\bottomrule
\end{tabular}
\caption{Passes included in the per-step FLOP estimate. All token counters sum over the relevant trajectories; rollout counts are not additional multipliers.}
\label{tab:flop-breakdown}
\end{table}

Both GRPO baselines use $G=8$ trajectories for policy training. Video-R1~\cite{video_r1} additionally samples $T=4$ trajectories to compute its temporal reward. These extra trajectories contribute to generation FLOPs (prefill and decode), but not to the policy update or reference-model KL pass. The teacher, contrastive, and reference passes are forward-only. Both GRPO baselines use the publicly available Video-R1 training pipeline, modified for vLLM colocation and updated to the same dependency versions as VIOS for fair comparison.

\section{Training and Generation Hyperparameters}
\label{app:hyperparams}

Table~\ref{tab:vios_training_hyperparameters} provides the full configuration used for VIOS training rollouts and optimization. The maximum completion length is model-dependent: $768$ tokens for Qwen2.5-VL~\cite{qwen25vl, video_r1} and $4096$ tokens for InternVL2.5~\cite{internvl25}.

\begin{table}[h]
\centering
\scriptsize
\setlength{\tabcolsep}{2.4pt}
\renewcommand{\arraystretch}{1.15}
\begin{tabular}{ll}
\toprule
\textbf{Hyperparameter} & \textbf{Value} \\
\midrule
Sampling temperature & $1.1$ \\
Top-$k$ & $20$ \\
Top-$p$ & $0.95$ \\
Training seed & $42$ \\
Per-device batch size & $8$ \\
Learning rate & $2 \times 10^{-6}$ \\
LR scheduler & Cosine (1-epoch horizon) \\
Weight decay & $0.01$ \\
Warmup ratio & $0.0$ \\
EMA decay $\alpha$ & $0.99$ (constant) \\
Contrastive weight $\lambda$ & $0.5$ \\
\bottomrule
\end{tabular}
\caption{VIOS training-rollout and optimization hyperparameters. Maximum completion length differs between Qwen2.5-VL and InternVL; all other settings are shared.}
\label{tab:vios_training_hyperparameters}
\end{table}

\paragraph{Training seed.}
The publicly released checkpoints of the video-reasoning baselines~\cite{video_r1, videochat_r1, video_r2, videorft, videochat_r15} were trained by their respective authors using seed $42$, the default in Hugging Face TRL, the framework on which our implementation is also based. For a consistent comparison, we likewise report VIOS trained using seed $42$.

\section{Contrastive Weight and EMA Decay Ablations}
\label{app:lambda_alpha_ablations}

\begin{table}[h]
\centering
\scriptsize
\setlength{\tabcolsep}{2.4pt}
\renewcommand{\arraystretch}{1.15}
\begin{tabular}{l BBB BBBB BB BB B B}
\toprule
& \multicolumn{3}{c}{\makebox[0pt][c]{\textbf{General}}}
  & \multicolumn{4}{c}{\makebox[0pt][c]{\textbf{Knowledge}}}
  & \multicolumn{2}{c}{\makebox[0pt][c]{\textbf{Temporal}}}
  & \multicolumn{2}{c}{\makebox[0pt][c]{\textbf{Long Video}}}
  & \textbf{Spatial}
  & \\
\cmidrule(lr){2-4} \cmidrule(lr){5-8} \cmidrule(lr){9-10} \cmidrule(lr){11-12} \cmidrule(lr){13-13}
\textbf{Variant}
& \rh{MVBench} & \rh{Video-MME} & \rh{MINERVA}
& \rh{MMVU} & \rh{Video-MMMU} & \rh{VideoMathQA} & \rh{SciVideoBench}
& \rh{TempCompass} & \rh{TVBench}
& \rh{LongVideoBench} & \rh{MLVU}
& \rh{VSI-Bench}
& \rh{Average} \\
\midrule
Qwen2.5-VL-SFT
& 60.72 & 55.15 & 29.55 & 59.36 & 47.22 & \underline{23.10} & 22.40 & 67.25 & 49.82 & 52.51 & 55.70 & 33.44 & 46.35 \\
$\lambda=0$
& 63.32 & 57.74 & 30.15 & 64.00 & 46.11 & \underline{23.10} & 24.70 & 70.93 & \textbf{52.55} & 54.23 & 61.09 & \textbf{36.41} & 48.69 \\
$\lambda=0.25$
& \underline{65.57} & \underline{58.85} & \underline{30.98} & \textbf{66.24} & \underline{48.00} & \underline{23.10} & \textbf{25.30} & 72.32 & 51.88 & 55.20 & 61.96 & 35.71 & \underline{49.59} \\
\rowcolor{rowgray}
$\lambda=0.5$ (Ours)
& 65.50 & \textbf{59.59} & \textbf{31.36} & \underline{65.76} & 47.44 & \textbf{25.48} & \underline{24.90} & \underline{72.77} & \underline{52.24} & \underline{55.87} & \textbf{62.74} & 35.73 & \textbf{49.95} \\
$\lambda=0.75$
& \textbf{65.72} & 58.15 & \underline{30.98} & 62.40 & \textbf{49.89} & \underline{23.10} & 23.40 & \textbf{73.82} & 52.16 & \textbf{56.54} & \underline{62.65} & \underline{36.23} & \underline{49.59} \\
$\lambda=1.0$
& 50.42 & 52.04 & 25.83 & 56.64 & 45.33 & 19.52 & 19.80 & 72.64 & 38.30 & 48.99 & 54.55 & 28.46 & 42.71 \\
\bottomrule
\end{tabular}
\caption{We sweep $\lambda$ while keeping all other training configurations identical. The $\lambda=0$ setting corresponds to the Distill. + Gate variant in Table~\ref{tab:ablation_loss}.}
\label{tab:ablation_contrastive_weight}
\end{table}

\paragraph{Contrastive Weight.} Between $\lambda = 0.25$ and 
$\lambda = 0.75$ the method is stable: the three averages lie within 0.4 points of one another, no benchmark moves by more than a few points, and no single setting dominates per-benchmark, so the reported result does not rest on a finely tuned $\lambda$. Performance declines substantially on average at $\lambda = 1.0$: once the contrastive term is weighted equally to distillation, divergence maximization dominates the objective and the student drifts from the teacher rather than being pulled toward it. The operative constraint is therefore that neither distillation nor contrasting should dominate, not a precise value of $\lambda$.

\begin{table}[h]
\centering
\scriptsize
\setlength{\tabcolsep}{2.4pt}
\renewcommand{\arraystretch}{1.15}
\begin{tabular}{l BBB BBBB BB BB B B}
\toprule
& \multicolumn{3}{c}{\makebox[0pt][c]{\textbf{General}}}
  & \multicolumn{4}{c}{\makebox[0pt][c]{\textbf{Knowledge}}}
  & \multicolumn{2}{c}{\makebox[0pt][c]{\textbf{Temporal}}}
  & \multicolumn{2}{c}{\makebox[0pt][c]{\textbf{Long Video}}}
  & \textbf{Spatial}
  & \\
\cmidrule(lr){2-4} \cmidrule(lr){5-8} \cmidrule(lr){9-10} \cmidrule(lr){11-12} \cmidrule(lr){13-13}
\textbf{Variant}
& \rh{MVBench} & \rh{Video-MME} & \rh{MINERVA}
& \rh{MMVU} & \rh{Video-MMMU} & \rh{VideoMathQA} & \rh{SciVideoBench}
& \rh{TempCompass} & \rh{TVBench}
& \rh{LongVideoBench} & \rh{MLVU}
& \rh{VSI-Bench}
& \rh{Average} \\
\midrule
Qwen2.5-VL-SFT
& 60.72 & 55.15 & 29.55 & 59.36 & 47.22 & 23.10 & 22.40 & 67.25 & 49.82 & 52.51 & 55.70 & 33.44 & 46.35 \\
$\alpha=0.90$
& \underline{65.62} & 58.56 & 30.23 & 64.00 & \underline{48.11} & 21.67 & \textbf{26.10} & \textbf{73.33} & \textbf{52.75} & \underline{56.32} & \underline{62.65} & \underline{35.22} & 49.55 \\
\rowcolor{rowgray}
$\alpha=0.99$ (Ours)
& 65.50 & \textbf{59.59} & \textbf{31.36} & \textbf{65.76} & 47.44 & \underline{25.48} & 24.90 & 72.77 & 52.24 & 55.87 & \textbf{62.74} & \textbf{35.73} & \textbf{49.95} \\
$\alpha=0.995$
& 65.50 & 59.00 & \underline{31.29} & 64.64 & 47.56 & 22.86 & 22.90 & 73.16 & \underline{52.71} & 55.95 & 61.96 & 33.64 & 49.26 \\
$\alpha=0.999$
& \textbf{65.75} & \underline{59.04} & 30.08 & 65.28 & \textbf{48.78} & \textbf{26.43} & \underline{25.70} & \underline{73.17} & 52.51 & \textbf{56.47} & 61.22 & 34.43 & \underline{49.91} \\
Fixed ($\alpha\to1$)
& 65.07 & 58.15 & 29.92 & \underline{65.60} & 47.78 & 23.81 & 24.10 & 72.92 & \underline{52.71} & 54.90 & 61.50 & 34.90 & 49.28 \\
\bottomrule
\end{tabular}
\caption{We sweep $\alpha$ while keeping all other training configurations identical. The $\alpha = 1.0$ setting corresponds to the Frozen Teacher variant in Table~\ref{tab:ablation_teacher}.}
\label{tab:ablation_ema_decay}
\end{table}

\paragraph{EMA decay.} All EMA settings improve the average performance substantially over the base model, with gains on nearly every benchmark. The choice of $\alpha$ matters less than the choice to update the teacher at all: the frozen teacher is not the best setting on any of the twelve benchmarks and is beaten on average by three of the four EMA values, while differences among the EMA settings themselves are mostly within one to two points per benchmark.

\section{Evaluation Protocol and Baseline Comparison}
\label{app:eval_protocol}

\paragraph{Unified evaluation protocol.}
We evaluate every model in Table~\ref{tab:video_benchmarks} ourselves under a single protocol. Specifically, we adopt the publicly released evaluation setup of Video-R1~\cite{video_r1} and VideoRFT~\cite{videorft} for all methods, including VIOS: at most 32 video frames, a per-frame pixel budget of $256\times28\times28$, temperature $0.01$, top-p $0.001$, and at most 1,024 generated tokens. All methods use the same prompt and evaluation pipeline, with InternVL-specific answer extraction. We evaluate the public checkpoints without retraining any baseline, so each retains the training configuration chosen by its authors. Separately, Table~\ref{tab:efficiency} matches Video-R1's per-step training configuration directly (Section~\ref{sec:efficiency}).

\paragraph{Changes to the reference implementation.}
Our evaluation differs from the original Video-R1/VideoRFT implementation in four respects. First, we update dependencies to their current versions: \texttt{vLLM} 0.7.x to 0.16.x, \texttt{torch} 2.5 to 2.10, and \texttt{transformers} 4.49 to 5.3. Second, we correct an issue in Video-R1's MVBench evaluator that silently omitted the video input for a subset of samples; the correction is applied to every model. Third, we preprocess benchmark videos into the model's input format before evaluation to reduce wall-clock time, without altering the inputs presented to the models or their outputs. Fourth, because InternVL2.5 models do not natively use \texttt{<answer>} tags, we add an InternVL-only fallback that extracts explicit final answers (e.g., ``the answer is B'' or ``Answer: (b)'') or, failing that, the last standalone uppercase option letter. The same extraction procedure is used for both InternVL base models and their VIOS checkpoints. The deviation from baselines' originally published numbers therefore stems from the baselines' own evaluation configurations, which differ substantially from one another and from our Video-R1/VideoRFT setup.

\section{Evaluation on Static Image Benchmarks}
\label{app:image_benchmarks}

To further assess the generality of VIOS across visual modalities, we evaluate the same VIOS-trained Qwen2.5-VL checkpoint on four static image understanding benchmarks. As shown in Table~\ref{tab:image_benchmarks}, VIOS improves average accuracy by 2.64 points over the SFT baseline. The largest gains occur on BLINK~\cite{blink} (+4.74) and CV-Bench~\cite{cvbench} (+4.47), which emphasize direct visual perception, while MMMU~\cite{mmmu} improves by 1.44 points, which is more knowledge-driven. ChartQA~\cite{chartqa} performance remains essentially unchanged (-0.08). These results demonstrate that VIOS's visual-reliance supervision generalizes beyond video to static image understanding.

\begin{table}[h]
\centering
\scriptsize
\setlength{\tabcolsep}{2.4pt}
\renewcommand{\arraystretch}{1.15}
\begin{tabular}{lccccc}
\toprule
\textbf{Model} & \textbf{ChartQA} & \textbf{BLINK} & \textbf{CV-Bench} & \textbf{MMMU} & \textbf{Average} \\
\midrule
Qwen2.5-VL-SFT & \textbf{84.64} & 51.18 & 71.76 & 51.78 & 64.84 \\
\rowcolor{rowgray}
+ VIOS (Ours) & 84.56 & \textbf{55.92} & \textbf{76.23} & \textbf{53.22} & \textbf{67.48} \\
\bottomrule
\end{tabular}
\caption{Evaluation on four static image understanding benchmarks. VIOS improves average accuracy by 2.64 points without image-specific modifications, demonstrating generalization beyond video reasoning.}
\label{tab:image_benchmarks}
\end{table}

\section{Training Stability and Seed Robustness}
\label{app:training_stability}

We assess VIOS training stability against vanilla GRPO across three seeds (42--44), using common diagnostics. Both methods start from Qwen2.5-VL-SFT, with VIOS trained for 100 steps and GRPO for 1{,}000 steps. During experimentation only standard diagnostics were logged for faster training. Gate-weight and entropy statistics were additionally logged for seeds 43 and 44.

\paragraph{Reliance-gap gate distribution.}
Table~\ref{tab:stability_gate} examines whether the gate progressively
concentrates or disperses its weights during training.

\begin{table}[h]
\centering
\scriptsize
\setlength{\tabcolsep}{4pt}
\renewcommand{\arraystretch}{1.13}
\begin{tabular}{lcccc}
\toprule
& \multicolumn{2}{c}{\textbf{Seed 43}} & \multicolumn{2}{c}{\textbf{Seed 44}} \\
\cmidrule(lr){2-3}\cmidrule(lr){4-5}
\textbf{Statistic} & \textbf{Start} & \textbf{End} & \textbf{Start} & \textbf{End} \\
\midrule
Mean         & 1.000 & 1.000 & 1.000 & 1.000 \\
Std. dev.    & 0.148 & 0.148 & 0.150 & 0.148 \\
$p_{05}$     & 0.774 & 0.742 & 0.772 & 0.741 \\
$p_{25}$     & 0.992 & 0.995 & 0.993 & 0.996 \\
$p_{50}$     & 1.025 & 1.034 & 1.025 & 1.034 \\
$p_{75}$     & 1.053 & 1.064 & 1.053 & 1.064 \\
$p_{95}$     & 1.114 & 1.114 & 1.117 & 1.113 \\
\bottomrule
\end{tabular}
\caption{Distribution of normalized VIOS gate weights at training start
and end across two seeds.}
\label{tab:stability_gate}
\end{table}

The trajectory-wise normalization (Eqs.~7--8) fixes the mean at one,
while the nearly unchanged standard deviation and quantiles show that
the gate neither sharpens nor flattens appreciably. Both seeds exhibit
nearly identical distributions, indicating stable token weighting
throughout VIOS training.

\paragraph{Gradient magnitudes and clipping.}
Table~\ref{tab:stability_grad} compares pre-clipping gradient norms under
the shared maximum gradient norm of 5.

\begin{table}[h]
\centering
\scriptsize
\setlength{\tabcolsep}{3.1pt}
\renewcommand{\arraystretch}{1.13}
\begin{tabular}{llrrrrrr}
\toprule
\textbf{Method} & \textbf{Seed} & \textbf{Mean} & \textbf{Std.} & \textbf{p95} & \textbf{p99} & \textbf{Max.} & \textbf{Clipped (\%)} \\
\midrule
GRPO & 42 & 2.45 & 3.90 & 4.76 & 10.8 & 112.4 & 4.4 \\
GRPO & 43 & 2.44 & 2.84 & 4.81 & 16.1 & 44.3 & 4.5 \\
GRPO & 44 & 2.46 & 2.85 & 4.84 & 10.9 & 57.2 & 4.7 \\
\midrule
\rowcolor{rowgray}
VIOS & 42 & 0.298 & 0.218 & 0.69 & 1.36 & 1.46 & 0.0 \\
\rowcolor{rowgray}
VIOS & 43 & 0.309 & 0.306 & 0.61 & 1.06 & 2.89 & 0.0 \\
\rowcolor{rowgray}
VIOS & 44 & 0.292 & 0.305 & 0.55 & 0.87 & 3.01 & 0.0 \\
\bottomrule
\end{tabular}
\caption{Pre-clipping gradient-norm statistics across three seeds.
Both methods use a clipping threshold of 5.}
\label{tab:stability_grad}
\end{table}

GRPO exhibits heavy-tailed gradients, with occasional large spikes
requiring clipping on approximately 4.5\% of steps. In contrast, VIOS
maintains substantially smaller and less variable gradients, with no
clipping across any seed. This indicates more controlled optimization
under VIOS.

\paragraph{KL drift and normalized token entropy.}
Table~\ref{tab:stability_policy} reports KL divergence from the base
policy and token entropy measured at temperature 1 for normalization.

\begin{table}[h]
\centering
\scriptsize
\setlength{\tabcolsep}{3.6pt}
\renewcommand{\arraystretch}{1.13}
\begin{tabular}{llccccc}
\toprule
& & & \multicolumn{4}{c}{\textbf{Normalized token entropy}} \\
\cmidrule(lr){4-7}
\textbf{Method} & \textbf{Seed} & \textbf{KL to base} & \textbf{Start} & \textbf{End} & $\boldsymbol{\Delta}$ & \textbf{Min.} \\
\midrule
GRPO & 42 & 0.0735 & -- & -- & -- & -- \\
GRPO & 43 & 0.0609 & 0.714 & 0.928 & $+0.214$ & 0.359 \\
GRPO & 44 & 0.0681 & 0.711 & 0.914 & $+0.203$ & 0.397 \\
\midrule
\rowcolor{rowgray}
VIOS & 42 & 0.0354 & -- & -- & -- & -- \\
\rowcolor{rowgray}
VIOS & 43 & 0.0349 & 0.721 & 0.694 & $-0.027$ & 0.620 \\
\rowcolor{rowgray}
VIOS & 44 & 0.0335 & 0.715 & 0.701 & $-0.014$ & 0.618 \\
\bottomrule
\end{tabular}
\caption{KL divergence from the base policy and normalized token
entropy. Entropy statistics are available for seeds 43 and 44;
dashes denote unavailable measurements.}
\label{tab:stability_policy}
\end{table}

GRPO undergoes larger changes in token uncertainty, reaching low
intermediate entropy before finishing substantially above its initial
level. VIOS shows only mild entropy decreases and consistently
maintains roughly half the KL divergence of GRPO. Together, these
results suggest more stable policy evolution under VIOS.

\paragraph{Completion lengths and homogeneous reward groups.}
Table~\ref{tab:stability_generation} compares generation lengths and,
for GRPO, the fraction of groups with identical rewards and hence
zero group-relative advantage.

\begin{table}[h]
\centering
\scriptsize
\setlength{\tabcolsep}{3.7pt}
\renewcommand{\arraystretch}{1.13}
\begin{tabular}{llrrrr}
\toprule
& & \multicolumn{3}{c}{\textbf{Completion length (tokens)}} & \textbf{Homog.} \\
\cmidrule(lr){3-5}
\textbf{Method} & \textbf{Seed} & \textbf{Start} & \textbf{End} & $\boldsymbol{\Delta}$ & \textbf{group frac.} \\
\midrule
GRPO & 42 & 290.6 & 376.5 & $+85.9$ & 0.299 \\
GRPO & 43 & 275.5 & 404.1 & $+128.6$ & 0.319 \\
GRPO & 44 & 282.9 & 384.3 & $+101.4$ & 0.292 \\
\midrule
\rowcolor{rowgray}
VIOS & 42 & 294.1 & 297.2 & $+3.1$ & -- \\
\rowcolor{rowgray}
VIOS & 43 & 295.1 & 293.9 & $-1.2$ & -- \\
\rowcolor{rowgray}
VIOS & 44 & 293.0 & 296.3 & $+3.3$ & -- \\
\bottomrule
\end{tabular}
\caption{Completion-length evolution and fraction of homogeneous-reward
GRPO groups across three seeds. Group homogeneity is not applicable to VIOS.}
\label{tab:stability_generation}
\end{table}

GRPO completions grow substantially, consistent with its explicit
length reward favoring 320--512 tokens, while approximately 30\% of
its sampled groups provide no group-relative learning signal.
VIOS avoids this group-estimation inefficiency and maintains nearly
unchanged completion lengths across all seeds, further demonstrating
consistent training behavior.

\section{Reliance-Gap Gate Limitations}
\label{app:gate_limitations}

Although the reliance-gap gate improves performance,
we identify a rare failure mode arising from noisy reliance
estimates at the final answer token. Occasionally, the gate
assigns a large weight to this position, amplifying both the
distillation and contrastive gradients.

For the contrastive term, the answer is already text-derivable
from the completed chain of thought, so the no-vision student
may already place substantial probability on the correct option.
Pushing the with-vision distribution away from its no-vision
counterpart can therefore perturb an already-correct prediction.
An analogous failure can arise in distillation: because the
EMA teacher is a lagged average of the student, it may disagree
with a student that has already improved onto the correct option,
pulling its prediction away from that answer.

While such perturbations can occur elsewhere, the answer token
is particularly fragile: its distribution is concentrated on
very few tokens, and its position at the end of the output makes
errors nonrecoverable. In rare cases, this produces a mismatch
between the answer reached by the chain of thought and the option
emitted between the answer tags.

\section{Additional Qualitative Results}

\begin{figure}[h]
    \centering
    \includegraphics[width=\textwidth]{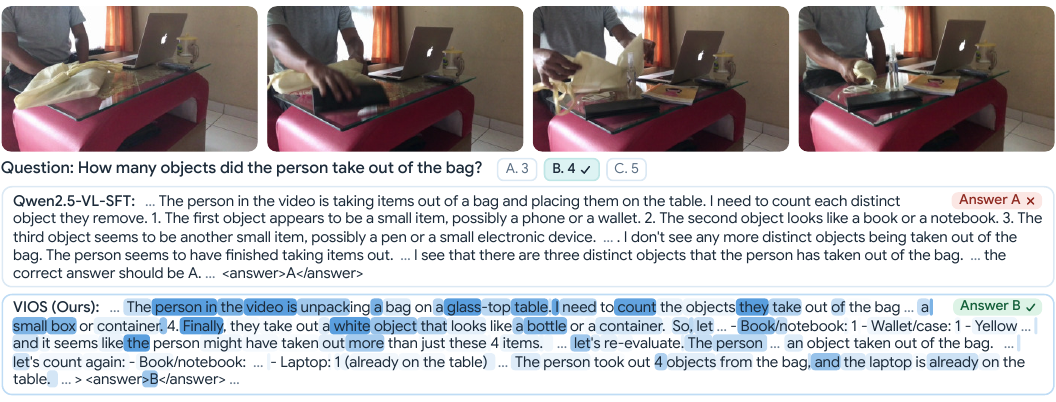}
    \includegraphics[width=\textwidth]{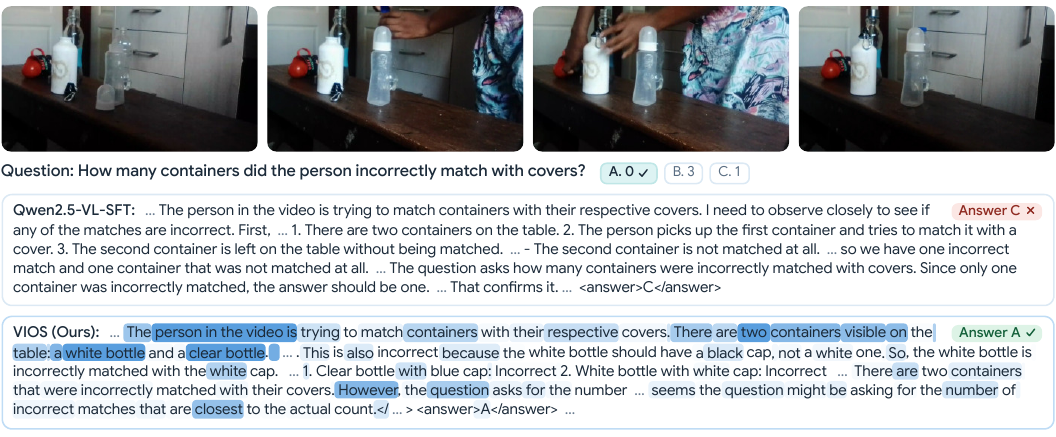}
    \includegraphics[width=\textwidth]{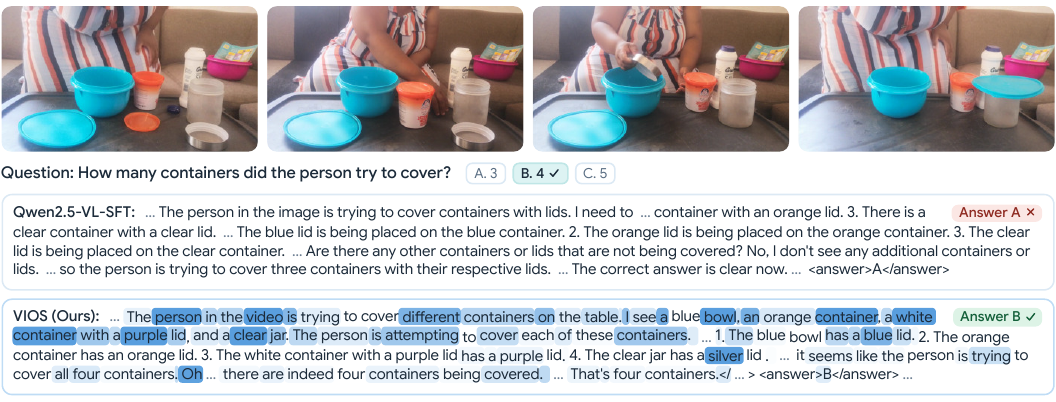}
    \caption{Additional reasoning traces from Qwen2.5-VL-SFT and VIOS on three video understanding tasks. VIOS tokens are shaded by per-token loss, with darker meaning higher loss.}
    \label{fig:qual_appendix}
\end{figure}

Figure~\ref{fig:qual_appendix} compares examples from Qwen2.5-VL-SFT and VIOS. In the top example, the base model stops at three objects, concluding ``I don't see any more distinct objects.'' The VIOS loss is highest on scene and object descriptors (``unpacking,'' ``glass-top table,'' ``small box,'' ``white object,'' ``bottle'') and on the counting cues ``Finally'' and ``more,'' where the model commits to an item or revises its tally. It then recounts, separates out the laptop that was ``already on the table,'' and settles on 4.

The middle example shows the loss exposing inconsistent reasoning. VIOS correctly identifies ``a white bottle'' and ``a clear bottle,'' both highly weighted, but then claims ``There are two containers that were incorrectly matched,'' which contradicts the ground truth and is not among the answer options. The loss is elevated on the visual claims behind this error (``white'' cap, ``black'' cap). Because the answer-privileged teacher would not produce this trajectory, VIOS flags the flawed step. Although the final answer (A) is correct, the signal points to the incorrect reasoning that training should correct.

In the bottom example, the base model sees only three containers and states, ``No, I don't see any additional containers or lids.'' VIOS names all four, and the loss falls on their distinguishing attributes (``bowl,'' ``white container,'' ``purple'' lid, ``clear jar,'' ``silver'' lid) and on the self-correction ``Oh'' just before the model confirms ``there are indeed four containers.'' Across all three examples, VIOS concentrates supervision on the few tokens that depend on the video, such as object identities, attributes, and count updates, rather than spreading it evenly over the chain of thought.

\section{Societal Impact}
\label{app:societal_impact}

VIOS reduces the compute required to post-train a competitive video reasoner by roughly $60\times$ in total training FLOPs and $111\times$ in tokens generated relative to the baselines (Section~\ref{sec:efficiency}), broadening access to research groups without large GPU budgets, lowering the energy footprint of post-training, and supporting beneficial downstream applications such as assistive video description, educational content understanding, and analysis of long-form media. These capabilities are dual-use, however, and the same improvements lower the cost of automated surveillance and large-scale video classification of individuals. VIOS also inherits the biases of its base model, since it is a post-training procedure that introduces no new data or modalities. By making strong video reasoning attainable at a fraction of the usual training cost, VIOS shifts the frontier toward broader and more sustainable participation in multimodal research.

\section{Prompt Templates}
\label{app:prompts}

\begin{figure}[ht]
    \centering
    \includegraphics[width=\textwidth]{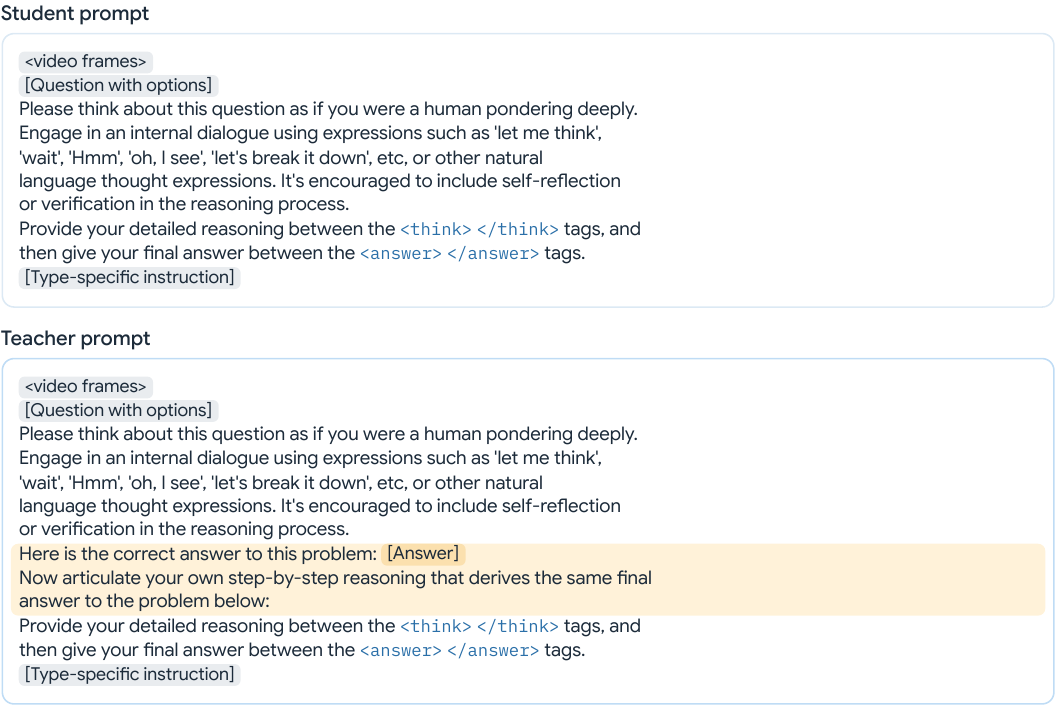}
    \caption{The privileged ground-truth answer is highlighted in the teacher prompt. Type-specific instruction refers to the question type. For example, for multiple-choice questions the instruction is to respond with either A, B, C, D, or E.}
\end{figure}

We utilize the same prompt template as Video-R1 for the student and include the highlighted section for the teacher prompt.